%% file: main.tex
\pdfoutput=1

\documentclass[10pt,twocolumn,letterpaper]{article}

\usepackage[pagenumbers]{cvpr} %

\usepackage{graphicx}
\usepackage{amsmath}
\usepackage{amssymb}
\usepackage{booktabs}
\usepackage{multirow}
\usepackage{adjustbox}

\usepackage[utf8]{inputenc} %
\usepackage[T1]{fontenc}    %
\usepackage{url}            %
\usepackage{booktabs}       %
\usepackage{amsfonts}       %
\usepackage{nicefrac}       %
\usepackage{microtype}      %
\usepackage{xcolor}         %
\usepackage{xspace}

\usepackage{bbm}

\usepackage[pagebackref,breaklinks,colorlinks]{hyperref}

\usepackage[capitalize]{cleveref}
\crefname{section}{Sec.}{Secs.}
\Crefname{section}{Section}{Sections}
\Crefname{table}{Table}{Tables}
\crefname{table}{Tab.}{Tabs.}

\makeatletter
\DeclareRobustCommand\onedot{\futurelet\@let@token\@onedot}
\def\@onedot{\ifx\@let@token.\else.\null\fi\xspace}

\def\eg{\emph{e.g}\onedot}

\makeatother

\newcommand{\E}{\mathsf{E}}

\newcommand{\system}{D-NeRV\xspace}

\def\cvprPaperID{2941} %
\def\confName{CVPR}
\def\confYear{2023}

\begin{document}

\title{Towards Scalable Neural Representation for Diverse Videos}

\author{Bo He$^{1}$\quad Xitong Yang$^{2}$\quad Hanyu Wang$^{1}$\quad Zuxuan Wu$^{3}$\quad Hao Chen$^{1}$\\ Shuaiyi Huang$^{1}$\quad Yixuan Ren$^{1}$\quad Ser-Nam Lim$^{2}$\quad Abhinav Shrivastava$^{1}$\\[0.5em]
$^{1}$University of Maryland, College Park \quad\quad $^{2}$Meta AI \quad\quad $^{3}$Fudan University\\
{\tt\small ~\url{https://boheumd.github.io/D-NeRV/}}
}
\maketitle

\input{sec/0_abstract}
\input{sec/1_introduction}

\input{sec/2_related}
\input{sec/3_method}

\input{sec/4_results}
\input{sec/5_conclusions}

\medskip
\noindent\textbf{Acknowledgements.} This project was partially funded an independent grant from Facebook AI.

{\small
\bibliography{egbib,main}
}
\clearpage

\input{appendix}

\end{document}

%% file: sec/0_abstract.tex
\begin{abstract}
Implicit neural representations (INR) have gained increasing attention in representing 3D scenes and images, and have been recently applied to encode videos (\eg, NeRV~\cite{chen2021nerv}, E-NeRV~\cite{li2022nerv}).
While achieving promising results, existing INR-based methods are limited to encoding a handful of short videos (\eg, seven 5-second videos in the UVG dataset) with redundant visual content, leading to a model design that fits individual video frames independently and is not efficiently scalable to a large number of diverse videos.
This paper focuses on developing neural representations for a more practical setup -- encoding long and/or a large number of videos with diverse visual content.
We first show that instead of dividing videos into small subsets and encoding them with separate models, encoding long and diverse videos jointly with a unified model achieves better compression results. Based on this observation, we propose \system, a novel neural representation framework designed to encode diverse videos by (i) decoupling clip-specific visual content from motion information, (ii) introducing temporal reasoning into the implicit neural network, and (iii) employing the task-oriented flow as intermediate output to reduce spatial redundancies.
Our new model largely surpasses NeRV and traditional video compression techniques on UCF101 and UVG datasets on the video compression task. Moreover, when used as an efficient data-loader, \system achieves 3\%-10\% higher accuracy than NeRV on action recognition tasks on the UCF101 dataset under the same compression ratios.
\end{abstract}

%% file: sec/1_introduction.tex
\vspace{-0.1in}
\section{Introduction}

\begin{figure}[t]
\centering
    \vspace{-0.1in}
    \adjincludegraphics[width=\linewidth, trim={{0.0\width} {0.0\height} {0.0\width} {0.0\height}},clip]{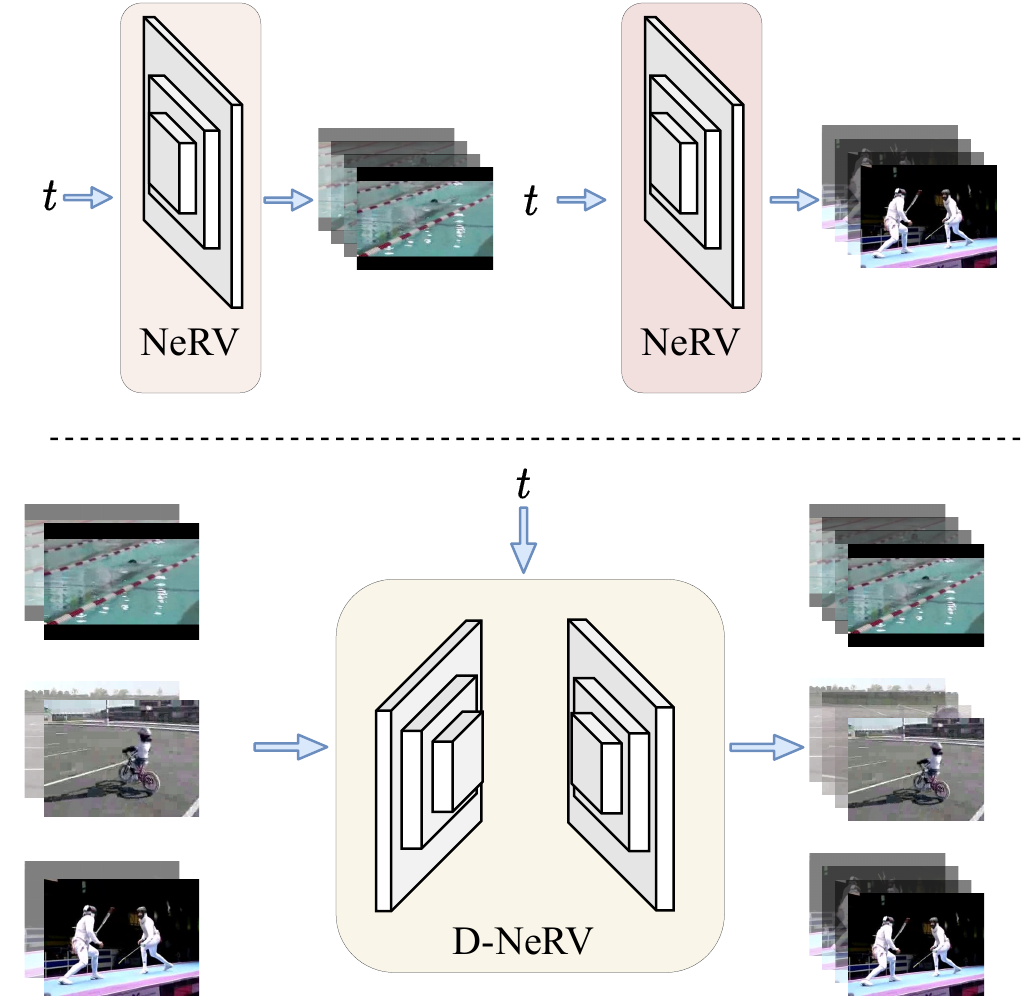}
    \caption{Comparison of \system and NeRV when representing diverse videos. NeRV optimizes representation to every video independently while \system encodes all videos by a shared model.}
    \label{fig:teaser}
\vspace{-0.2in}
\end{figure}

Implicit neural representations (INR) have achieved great success in parameterizing various signals, such as 3D scenes~\cite{mildenhall2020nerf,yu2021pixelnerf,sitzmann2019scene}, images~\cite{sitzmann2020implicit,chen2021learning}, audio~\cite{sitzmann2020implicit}, and videos~\cite{chen2021nerv,li2022nerv,chen2022cnerv,maiya2023nirvana,chen2023hnerv}.
The key idea is to represent signals as a function approximated by a neural network, mapping a reference coordinate to its corresponding signal value.
Recently, INR has received increasing attention in image and video compression tasks~\cite{dupont2021coin,chen2021nerv,li2022nerv,rho2022neural,zhang2021implicit,dupont2022coin++,strumpler2021implicit,chen2022cnerv}.
Compared with learning-based video compression techniques~\cite{lu2019dvc,yang2020Learning,li2021deep}, INR-based methods (\eg, NeRV~\cite{chen2021nerv}) are more favorable due to simpler training pipelines and much faster video decoding speed.

\begin{figure}[t]
\centering
    \vspace{-0.15in}
    \adjincludegraphics[width=\linewidth, trim={{0.0\width} {0.02\height} {0.0\width} {0.0\height}},clip]{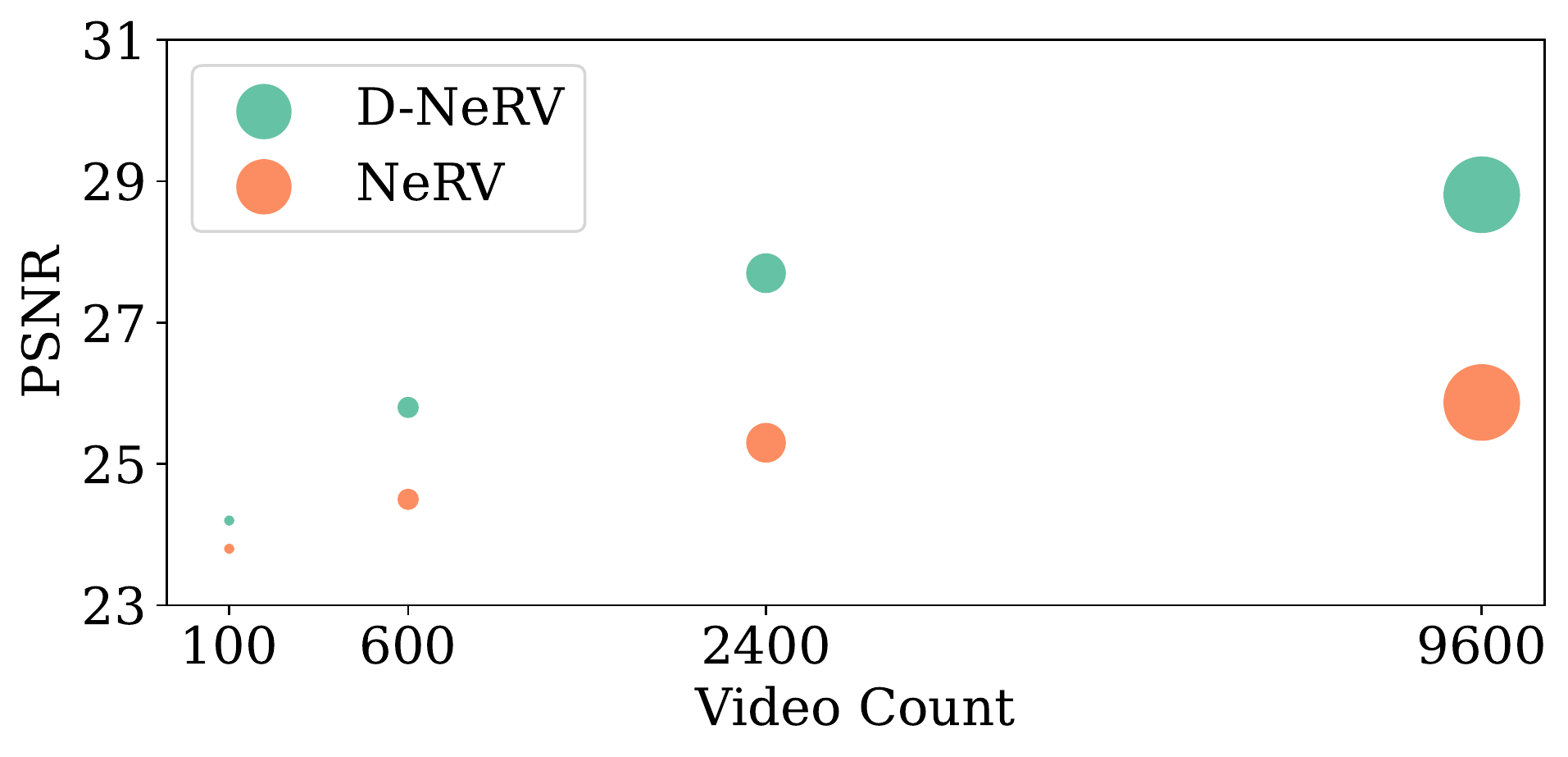}
    \caption{Comparison of \system and NeRV with fixed compression ratio on UCF101. The size of circles indicates model sizes.}
    \label{fig:teaser_fixbpp}
\vspace{-0.15in}
\end{figure}

While impressive progress has been made, existing INR-based methods are limited to encoding a single short video at a time.
This prohibits the potential applications in most real-world scenarios, where we need to represent and compress a large number of diverse videos.
A straightforward strategy for encoding diverse videos is to divide them into multiple subsets and model each of them by a separate neural network, as shown in Figure~\ref{fig:teaser} (top).
However, since this strategy is unable to leverage long-term redundancies across videos, it achieves inferior results compared to fitting all diverse videos with a single shared model.
As shown in Figure~\ref{fig:teaser_fixbpp}, under the same compression ratio (bits per pixel), the performance of NeRV is consistently better when fitting a larger number of videos.
This suggests that representing multiple videos by a single large model is generally more beneficial.

However, as observed empirically, the current design of NeRV offers diminishing returns when scaling to large and diverse videos. 
We argue that the current coupled design of content and motion information modeling exaggerates the difficulty of memorizing diverse videos.
To address this, we propose \system, a novel implicit neural representation that is specifically designed to efficiently encode long or a large number of diverse videos\footnote{``Long videos" and ``a large number of videos" are viewed as interchangeable concepts in this paper because a long video can be obtained by concatenating a collection of diverse videos.}.
A representative overview of differences between \system and NeRV is shown in Figure~\ref{fig:teaser}.
When representing diverse videos, NeRV encodes each video into a separate model or simply concatenates multiple videos into a long video and encodes it, while our \system can represent different videos in a single model by conditioning on key-frames from each video clip.

Compared to NeRV, we have the following improvements. First, we observe that the visual content of  each video often represents appearance, both background and foreground, which vary significantly among different videos, while the motion information often represents the semantic structure (\eg, similar motion for the same action class) and can be shared across different videos.
Therefore, we decouple each video clip into two parts: clip-specific visual content and motion information, which are modeled separately in our method. 
Second, motivated by the vital importance of temporal modeling in video-related tasks, instead of outputting each frame independently, we introduce temporal reasoning into the INR-based network by explicitly modeling global temporal dependencies across different frames.
Finally, considering the significant spatial redundancies in videos, rather than predicting the raw pixel values directly, we propose to predict the task-oriented flow~\cite{xue2019video,reda2022film,huang2019dynamic,huang2022learning} as an intermediate output, and use it in conjunction with the key-frames to get the final refined output. It alleviates the complexity of memorizing the same pixel value across different frames.

With these improvements, our \system significantly outperforms NeRV, especially when increasing the number of videos as shown in Figure~\ref{fig:teaser_fixbpp}. 
To summarize, our main contributions are as follows:
\begin{itemize}
    \vspace{-0.5em}
    \item We propose \system, a novel implicit neural representation model, to represent a large and diverse set of videos as a single neural network.
    \vspace{-0.5em}
    \item We conduct extensive experiments on video reconstruction and video compression tasks. Our \system consistently outperforms state-of-the-art INR-based methods (E-NeRV~\cite{li2022nerv}), traditional video compression approaches (H.264~\cite{wiegand2003overview},HEVC~\cite{sullivan2012overview}), and the recent learning-based video compression methods (DCVC~\cite{li2021deep}).
    \vspace{-0.5em}
    \item We further show the advantage of \system on the action recognition task by its higher accuracy and faster decoding speed, and reveal its intriguing properties on the video inpainting task.
\end{itemize}

%% file: sec/2_related.tex
\section{Related Work}

\noindent\textbf{Implicit neural representations.}
Neural networks can be used to approximate the functions which maps the input coordinates to various types of signals.
It has brought great interest and has been widely adopted to represent 3D shape ~\cite{sitzmann2019scene,mescheder2019occupancy}, novel view synthesis~\cite{mildenhall2020nerf,yu2021pixelnerf}.
These approaches train a neural network to fit a single scene or object such that it is encoded by the network weights.
Implicit neural representations have also been applied to represent images ~\cite{sitzmann2020implicit,chen2021learning,dupont2021coin}, videos~\cite{chen2021nerv,rho2022neural} and audios~\cite{sitzmann2020implicit}.
Among them, NeRV proposes the first image-wise implicit neural representation for videos, which takes the frame index and outputs the corresponding RGB frame. 
Compared to the pixel-wise implicit neural representation SIREN~\cite{sitzmann2020implicit}, NeRV shows superior efficiency, which improves the encoding and decoding speed greatly and achieves better video reconstruction quality.
Based on NeRV, E-NeRV~\cite{li2022nerv} boosts the video reconstruction performance via decomposing the image-wise implicit neural representation into separate spatial and temporal contexts.
NRFF~\cite{rho2022neural} and IPF~\cite{zhang2021implicit} predict the motion compensation and residual between consecutive video frames to better reduce the spatial redundancies.
CNeRV~\cite{chen2022cnerv} proposes a hybrid video neural representation with content-adaptive embedding to further introduce internal generalization.

\begin{figure*}[t!]
    \centering
    \vspace{-0.15in}
    \includegraphics[width=\textwidth]{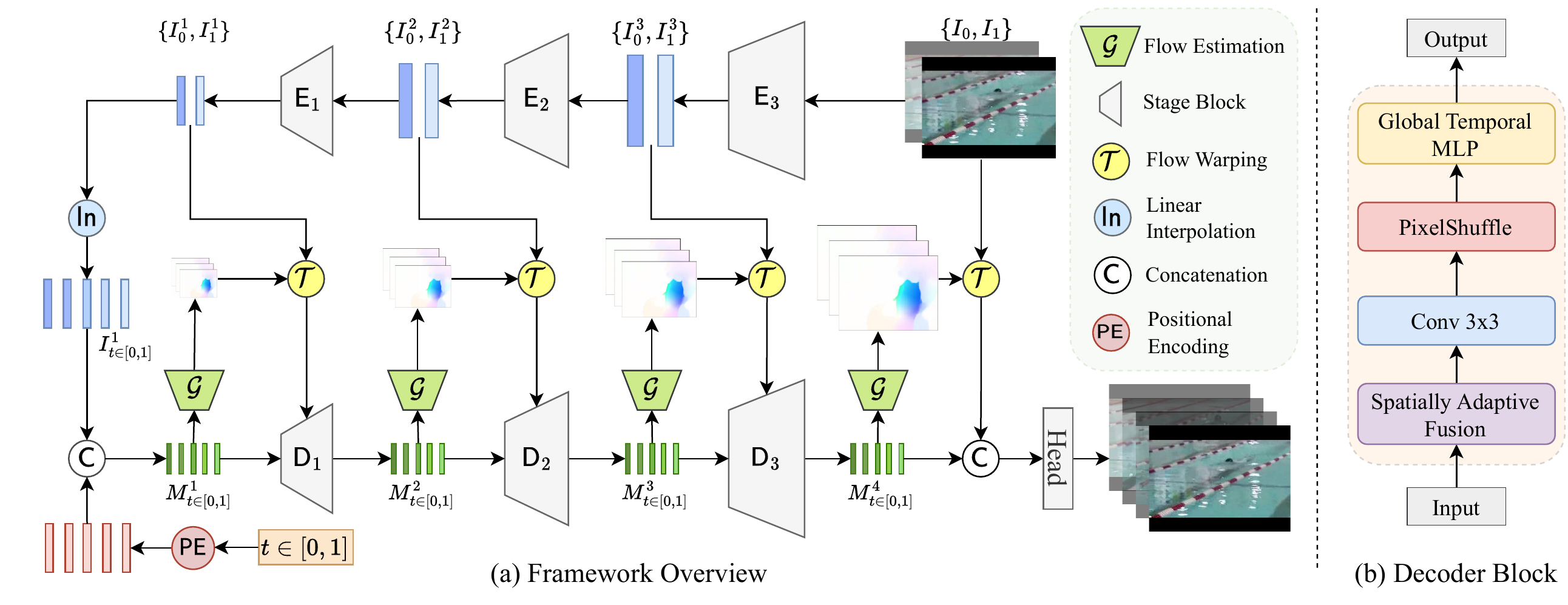}
    \vspace{-0.25in}
    \caption{(a) Frame Overview. \system takes in key-frame pairs of each video clip along with all the frame indices and outputs a whole video clip at a time. (b) The decoder block predicts the flow estimation to warp the visual content feature from the encoder, then fuses the visual content by the spatially-adaptive fusion module and finally models temporal relationship by the global temporal MLP module.}
    \label{fig:model}
    \vspace{-0.1in}
\end{figure*}

\vspace{0.05in}
\noindent\textbf{Video Compression.}
Video compression techniques can be divided into traditional video compression algorithms, such as MPEG~\cite{le1991mpeg}, H.264~\cite{wiegand2003overview}, HEVC~\cite{sullivan2012overview} and deep learning-based compression approaches, such as DVC~\cite{lu2019dvc}, LVC~\cite{rippel2019learned}, HLVC~\cite{yang2020Learning}, DCVC~\cite{li2021deep}.
The learning-based compression approaches often use convolutional neural networks (CNN) to replace certain components while still following the traditional video compression pipeline.
Recently, INR-based models have been adopted to compress image and video data. 
They encode images and videos into neural networks and apply general model compression techniques, which convert the image and video compression task to a standard model compression task.
COIN~\cite{dupont2021coin} uses a vanilla pixel-wise INR model to fit images and adopts weight quantization to do model compression. 
NeRV~\cite{chen2021nerv} applies model pruning, weight quantization, and entropy encoding to further reduce the model size.

Among these INR-based methods, NeRV~\cite{chen2021nerv} is the first image-wise INR-based model specifically designed for videos. 
As is shown in Figure~\ref{fig:teaser}, NeRV takes in the time index $t$ as input and outputs the corresponding frame directly, which can represent a video as a neural network.
Specifically, NeRV consists of a positional encoding function, stacked NeRV blocks, and a prediction head. More details can be found in NeRV paper~\cite{chen2021nerv}.
Although with such a simple architecture, NeRV shows promising results in the video compression task.
However, for large-scale and diverse videos, NeRV encodes each video into a separate model or simply concatenates them into one longer video.
This design is not optimal when modeling such massive information into a neural network, which motivates us to design a more effective framework for diverse video encoding.

%% file: sec/3_method.tex
\vspace{-0.05in}
\section{Method}
\vspace{-0.05in}

To effectively represent diverse videos by a single model, we propose \system.
Figure~\ref{fig:model}(a) illustrates the overview of \system framework.
Given each video clip, we decouple the clip-specific visual content from the motion information and model each of them by two main components of our \system.
Specifically, we introduce a visual content encoder to encode the clip-specific visual content from the sampled key-frames and a motion-aware decoder to output video frames.
We elaborate on the details of the visual content encoder (Sec.~\ref{sec:encoder}), the motion-aware decoder (Sec.~\ref{sec:decoder}), and the training process (Sec.~\ref{sec:train}) in the following sections.

\vspace{-0.02in}
\subsection{Visual Content Encoder}
\label{sec:encoder}
Different videos have various content information, \eg, the appearance and the background scene of each video vary greatly. 
The first component of \system is a visual content encoder $\E$ to capture clip-specific visual content. 
In contrast to existing works which memorize the content of diverse videos solely by the model itself~\cite{chen2021nerv,li2022nerv,rho2022neural,zhang2021implicit}, we propose to provide the network with the visual content via sampled key-frames.
Intuitively, we divide each video into consecutive clips. For each video clip, we sample the start and end key-frames $(I_0, I_1)$, which are then fed into the content encoder $\E$ to extract visual content at multiple stages $\{I^l_0, I^l_1\}_{l=1}^{L} = \{\E(I_0), \E(I_1)\}$ ($L$ is the total number of stages).
These extracted features are clip-specific and highly representative of video content.
Specifically, the content encoder $\E$ consists of stacked convolution layers and gradually down-samples the key-frames.

\subsection{Motion-aware Decoder}
\label{sec:decoder}
Although different videos have distinctive appearances or backgrounds, videos of the same action type can share similar motion information.
Motivated by this observation, we propose to model the motion information by a shared implicit neural network based decoder. 
With visual content from key-frames, the motion-aware decoder provides motion information to reconstruct the full video.
While the standard implicit neural network only takes in the coordinates and outputs the corresponding signal values~\cite{sitzmann2020implicit,dupont2021coin,dupont2022coin++,chen2021nerv}, our motion-aware decoder takes in both the time coordinates and the content feature map. Then it predicts task-oriented flows as intermediate output, which are used to warp the generated content features. 
Besides that, we propose a spatially-adaptive fusion module to fuse the content information into the decoder in a more effective manner.
Finally, we equip the decoder with temporal modeling ability by the proposed global temporal MLP module.

\vspace{0.05in}
\noindent\textbf{Multi-scale Flow Estimation.}
The first component is the multi-scale flow estimation network used for predicting the task-oriented flow at each time step. 
At the first stage, given two content feature maps $I^1_0$ and $I^1_1$ from the encoder's output, we apply linear interpolation along the time axis to generate the feature map at every intermediate time step $I^1_t=\texttt{Interpolation}(I^1_0, I^1_1)(t)$.
Then, following NeRV~\cite{chen2021nerv}, we map the input time index $t$ by the positional encoding function $\mathsf{PE}$ into a higher dimensional embedding space, which is then concatenated with the feature map $I^1_t$ before fed into the flow estimation module $\mathcal{G}$ at the first stage:
\begin{equation}
    M^1_t = \texttt{Concat}(I^1_t, \mathsf{PE}(t))
\end{equation}
Next, we compute the forward flow $F_{t\rightarrow 0}$ and backward flow $F_{t\rightarrow 1}$ at each time $t\in [0,1]$ simultaneously, where $F_{t\rightarrow 0}$ and $F_{t\rightarrow 1}$ represent the pixel displacement map from the current frame $t$ to the start and end key-frames.
For the later stages, the input of the flow estimation module is the feature map $M^l_t$ generated from the previous decoder stage:
\begin{equation}
    F^l_{t\rightarrow 0}, F^l_{t\rightarrow 1} = \mathcal{G}^l(M^l_t), l \in \{1, \cdots, L\}
\end{equation}
where $\mathcal{G}$ is a stack of convolutions which calculates the per-pixel flow.
In order to generate a high-quality content feature map for each timestep, we strategically propagate the visual content of key-frames \{$I^l_0,I^l_1$\} to the current frame index $t$ under the guidance of our estimated flows \{$F^l_{t\rightarrow 0}, F^l_{t\rightarrow 1}$\}. 
Concretely, we first generate the forward (resp. backward) warped feature map $\hat{I}^l_{t\leftarrow 0}$ (resp. $\hat{I}^l_{t\leftarrow 1}$) at time index $t$ given the content features of key frame $I^l_0$ (resp. $I^l_1$) with its corresponding flow $F^l_{t\rightarrow 0}$ ($F^l_{t\rightarrow 1}$) by a bilinear warp operation $\mathcal{T}$:
\begin{equation}
    \hat{I}^l_{t\leftarrow 0} = \mathcal{T}\left(I^l_0, F^l_{t\rightarrow 0}\right), \hat{I}^l_{t\leftarrow 1} = \mathcal{T}\left(I^l_1, F^l_{t\rightarrow 1}\right)
\end{equation}
To fuse the forward and backward warped feature map in a reliable way, we devise a distance-aware confidence score to weighted sum the warped features and generate the fused warping feature $\hat{I}^l_t$:
\begin{align}
     \hat{I}^l_t &= (1-t) \cdot \hat{I}^l_{t\rightarrow 0} + t \cdot \hat{I}^l_{t\rightarrow 1}
\end{align}

\vspace{0.05in}
\noindent\textbf{Spatially-adaptive Fusion (SAF). }
The warped feature map $\hat{I}^l_t \in \mathbb{R}^{H^l\times W^l\times C}$ contains the clip-specific content information for each timestep.
We further introduce the second module of our motion-aware decoder, a spatially-adaptive fusion module to fuse the clip-specific content information.
This is motivated by the recent success of modulation layers~\cite{park2019semantic,huang2017arbitrary,karras2019style}.
Specifically, we learn pixel-wise modulation parameters $\gamma^l_t, \beta^l_t$ by passing the content feature $\hat{I}^l_t$ into two fully-connected layers:
\begin{equation}
    \gamma^l_t = FC_1(\hat{I}^l_t), \beta^l_t = FC_2(\hat{I}^l_t)
\end{equation}
where $\gamma^l_t, \beta^l_t \in \mathbb{R}^{H^l\times W^l\times 1}$. Then we fuse $M^l_t$ as follows:
\begin{equation}
    J^l_t = \gamma^l_t M^l_t + \beta^l_t
\end{equation}
It introduces an additional inductive bias guided by the content feature $\hat{I}^l_t$, which integrates two feature maps in a more effective way than simple concatenation. 
After the modulation operation, we adopt the same block architecture as NeRV~\cite{chen2021nerv}, which consists of one convolution layer, a GELU~\cite{hendrycks2016gaussian} activation layer, and a PixelShuffle layer~\cite{shi2016real} to gradually upsample the feature map as below:
\begin{equation} 
    O^l_t = \texttt{PixelShuffle}\left(\texttt{GELU}(\texttt{Conv}(J^l_t))\right)
\end{equation}

\vspace{0.05in}
\noindent\textbf{Global Temporal MLP (GTMLP). }
Recall that NeRV takes the time index as input and outputs the corresponding frame directly without considering the rich intrinsic temporal correlations across frames. 
Inspired by the recent success of attention-based transformers~\cite{bertasius2021space,fan2021multiscale} and MLP-based models~\cite{tolstikhin2021mlp,touvron2021resmlp,he2020gta,lian2021mlp} in image and video recognition tasks, we introduce a global temporal MLP module to further exploit the temporal relationship of videos.
Compared to transformers, MLP-based models are more lightweight and efficient, which only consist of highly optimized fully-connected layers.
Motivated by this, we propose a global temporal MLP module to model the temporal relationship across different frames.
Specifically, given the feature map of $T$ frames, $O^l \in \mathbb{R}^{C \times H\times W\times T}$, the fully connected layer with weight $W^l \in \mathbb{R}^{C \times T \times T}$ is applied for each channel along the time axis to model the global temporal dependencies, which then adds with the original feature map $O^l$ in a residual manner.
\begin{equation}
    M^{l+1} = O^l + \texttt{matmul}(O^l, W^l)
\end{equation}

\noindent\textbf{Final Stage.}
To generate the final reconstructed frame $I'_t$ at time index $t$, we concatenate the decoder feature map $M^L_t$, the warped frame $\hat{I}_t$ as input, feeding it into a stack of two convolution layers for the final refinement.

\subsection{Training}
\label{sec:train}
\vspace{-0.02in}

We adopt a combination of L1 and SSIM loss as~\cite{wang2003multiscale} between the reconstructed frame $I'_t$ and the ground-truth frame $I_t$ for optimization same as NeRV, without any explicit supervision on flow estimation. 
\begin{equation}
    \mathcal{L}=\dfrac{1}{T} \sum_t \left\lVert I'_t - I_t \right\rVert_1 + \alpha \left(1 - \text{SSIM}(I'_t, I_t)\right)
\end{equation}
During training, we feed consecutive video clips over the entire dataset in a mini-batch manner and encode the selected key-frames with existing image compression algorithms.

Once trained, each video clip can be reconstructed by feeding time indices and clip-specific key-frames into \system. Our decoupled model design and novel learning strategy open up possibilities for large-scale video training, which fundamentally differs from the existing INR-based work that either fits videos into separate models or requires concatenating all videos as input with one model.

%% file: sec/4_results.tex
\vspace{-0.05in}
\section{Experiments}

\subsection{Setup}
\noindent\textbf{Datasets.} We evaluate our model on one widely used video action recognition dataset UCF101~\cite{soomro2012ucf101}, one standard video compression dataset UVG~\cite{mercat2020uvg} and DAVIS~\cite{perazzi2016benchmark} dataset for the video inpainting task.
\textbf{UCF101} contains 13320 videos of 101 different action classes and has a large diversity of action types with the presence of large variations in video content.
We extract all videos at 2 fps with 256$\times$320 spatial resolution. We follow the first training/testing split. 
\textbf{UVG} consists of 7 videos and 3900 frames in total. To compare with other learning-based video compression methods~\cite{lu2019dvc,li2021deep}, we also crop the UVG videos to 1024x1920. 
We take 10 videos from \textbf{DAVIS} validation split and crop them to the spatial size of 384x768.

\vspace{0.05in}
\noindent\textbf{Evaluation Tasks.}
To understand the video representation capability of different INR-based methods, we compare with SOTA INR-based methods (NeRV~\cite{chen2021nerv}, E-NeRV~\cite{li2022nerv}) on the task of video reconstruction in Section~\ref{sec:sotainr}.
Next, as video compression is considered as the most promising downstream application of INR-based video representations, we also validate the effectiveness of our \system on the UVG and UCF101 datasets in Section~\ref{sec:compression}. 
Furthermore, since our \system can represent large-scale and diverse videos in a single model, we naturally extend the application of our \system to use it as an efficient dataloader, and demonstrate its effectiveness on the downstream video understanding tasks~\cite{lin2019tsm,feichtenhofer2019slowfast,yang2021beyond,li20212d,saini2022recognizing,he2022asm,wang2022efficient,wang2022bevt,he2023a2summ} (\eg, action recognition) in Section~\ref{sec:recognition}.
Finally, we show intriguing properties and advantages of \system on the video inpainting task in Section~\ref{sec:inpainting}.\looseness=-1

\vspace{0.05in}
\noindent\textbf{Implementation Details.} In our ablation experiments, we train \system using the AdamW~\cite{loshchilov2017decoupled} optimizer.
We use the cosine annealing learning rate schedule, the batch size of 32, the learning rate of 5e-4, training epochs of 800 and 400, and warmup epochs of 160 and 80 for UCF101 and UVG datasets, respectively.
The key-frames for each video are sampled at stride 8 on both datasets.
Following~\cite{li2021deep}, we compress key-frames by using the image compression technique~\cite{cheng2020learned}.
When comparing with other implicit neural representations such as NeRV and E-NeRV, we sum the compressed key-frame size and model size as the total size for \system, and keep the total size of \system equal to the model size of NeRV and E-NeRV for a fair comparison.
The total sizes of different model variants (S/M/L) on the UCF101 dataset are 79.2/94.5/114.5 MB respectively.
More dataset-specific training and testing details are available in the supplementary material.

\begin{table*}[t]
\centering
\begin{minipage}{.72\textwidth}
    \centering
    \caption{Video reconstruction comparison between our \system, NeRV~\cite{chen2021nerv} and E-NeRV~\cite{li2022nerv} on 7 videos from the UVG dataset. We keep the total size of \system including the key-frame size and model size to be the same as the model size of NeRV and E-NeRV for a fair comparison. We report the PSNR results for each video. NeRV and E-NeRV are trained in separate models for each video while NeRV$^*$ and \system fit multiple videos in a shared model.}
    \resizebox{0.988\linewidth}{!}{
    \renewcommand{\arraystretch}{1.25}
    \begin{tabular}{@{}l|ccccccc|c@{}} 
        \toprule
        Video & Beauty & Bosphorus & Bee & Jockey & SetGo & Shake & Yacht & avg.\\
        \midrule
        NeRV & 33.06 & 32.38 & 37.88 & 31.18 & 24.02 & 33.48 & 26.91 & 31.27 \\
        E-NeRV & 33.07 & 33.52 & 39.36 & 30.88 & 25.19 & 34.6 & 28.21 & 32.12 \\
        \midrule 
        NeRV$^*$ & 32.71 & 33.36 & 36.74 & 32.16 & 26.93 & 32.69 & 28.48 & 31.87 \\
        D-NeRV &  33.77 & 38.66 & 37.97 & 35.51 & 33.93 & 35.04 & 33.73 & \bf 35.52 \\
        \bottomrule
    \end{tabular}
    }
    \label{tab:enerv}
\end{minipage}
\hfill
\begin{minipage}{.24\textwidth}
    \caption{Video compression results on the UCF-101 dataset.}
    \resizebox{\linewidth}{!}{
    \renewcommand{\arraystretch}{1.1}
    \begin{tabular}{@{}l|cc@{}} 
        \toprule
        \textbf{Method} & PSNR & MS-SSIM \\
        \midrule
        H.264-S & 26.29 & 0.903 \\
        NeRV-S & 26.79 & 0.910 \\
        \system-S & \bf 28.15 & \bf 0.916 \\
        \midrule
        H.264-M & 27.42 & 0.925 \\
        NeRV-M & 27.35 & 0.921 \\
        \system-M & \bf 29.18 & \bf 0.937 \\
        \midrule
        H.264-L & 28.54 & 0.941 \\
        NeRV-L & 27.57 & 0.928 \\
        \system-L & \bf 30.06 & \bf 0.951 \\
        \bottomrule
    \end{tabular}
    }
    \label{tab:ucf_compression}
\end{minipage}
\end{table*}

\begin{figure*}[t]
\vspace{-0.05in}
    \centering
    \subfloat[PSNR vs. BPP]{
        \includegraphics[width=.44\linewidth]{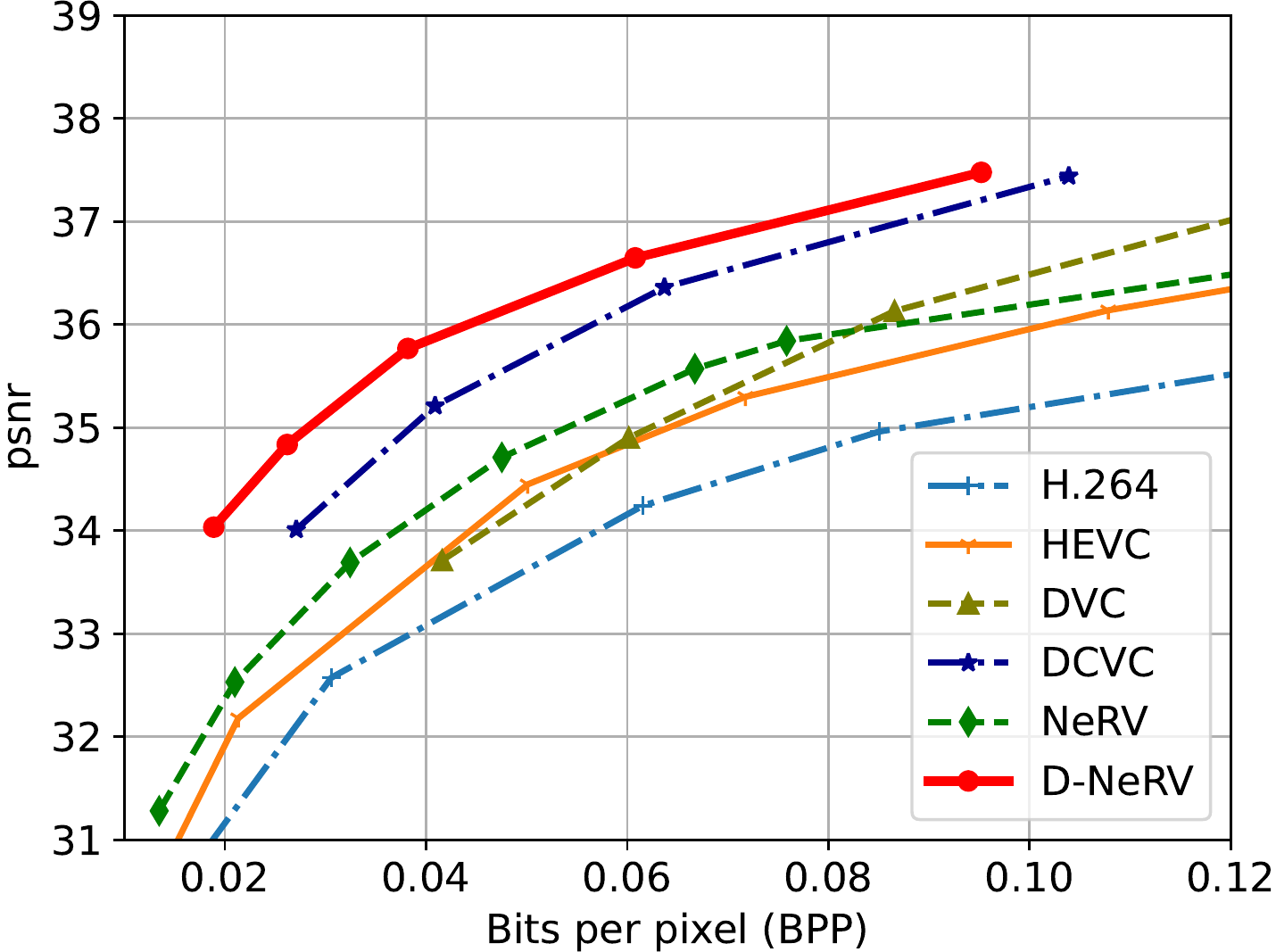}
        \label{fig:uvg_psnr}
    } 
    \hfill
    \subfloat[MS-SSIM vs. BPP ]{
        \includegraphics[width=.44\linewidth]{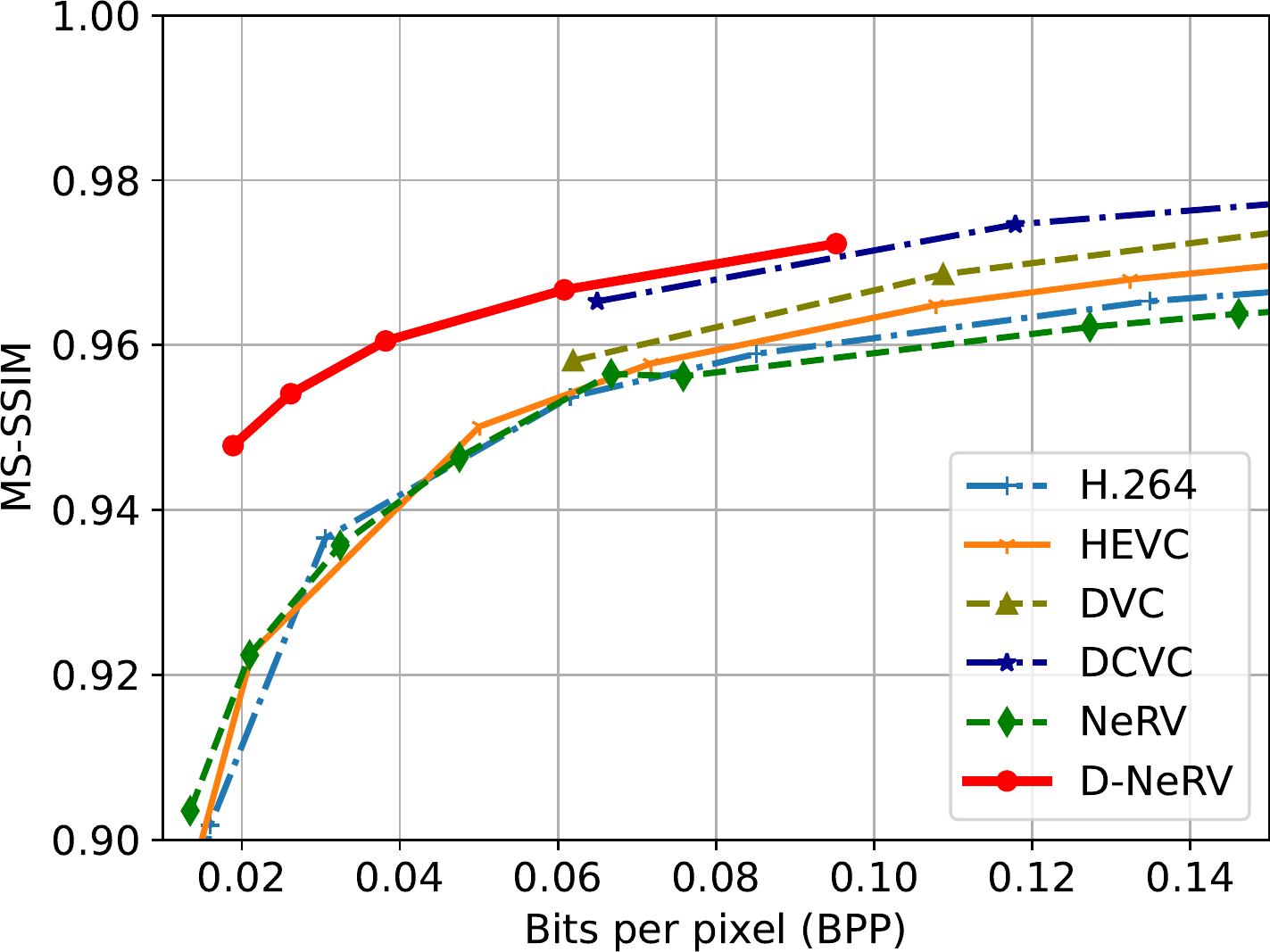}
        \label{fig:uvg_ssim}
    }
    \vspace{-0.1in}
    \caption{ Rate distortion plots on the UVG dataset. }
    \label{fig:uvg_results}
\vspace{-0.1in}
\end{figure*}

\subsection{Comparison with SOTA INRs}
\label{sec:sotainr}
We compare our \system with NeRV~\cite{chen2021nerv} and E-NeRV\cite{li2022nerv} on the UVG dataset for the \textbf{video reconstruction} task (without any compression steps). 
E-NeRV is the state-of-the-art INR-based video representation model.
The results are shown in Table~\ref{tab:enerv}, our \system can consistently outperform NeRV and E-NeRV on different videos of the UVG dataset. 
We first note that E-NeRV surpasses NeRV by 0.9 dB PSNR with the same model size.
As we mentioned before, encoding all the videos jointly with a shared model achieve better compression results (NeRV$^{*}$), which brings about a 0.6 dB performance boost than fitting each video along (NeRV).
Despite that, our \system achieves the best performance among these methods. Specifically, it outperforms the previous state-of-the-art INR-based method E-NeRV by 3.4 dB for the averaged PSNR.

\subsection{Video Compression}
\label{sec:compression}
We further evaluate the effectiveness of \system on the video compression task.
For video compression, we follow the same practice as NeRV for model quantization and entropy encoding but without model pruning to expedite the training process.

\vspace{0.05in}
\noindent\textbf{UCF101 Dataset.}
To demonstrate the effectiveness of \system in representing large-scale and diverse videos, in Table~\ref{tab:ucf_compression}, we show the comparison results of \system with NeRV~\cite{chen2021nerv} and H.264~\cite{wiegand2003overview} on the UCF101 dataset.
First, we observe that \system vastly outperforms the baseline model NeRV.
Especially when changing the model size from small (S) to large (L), the gap between \system and NeRV becomes larger, increasing from 1.4 dB to 2.5 dB.
It demonstrates that \system is more capable of compressing large-scale videos with high quality than NeRV.
Also, we can see that \system consistently surpasses traditional video compression techniques H.264, showing its great potential in real-world large-scale video compression.

\vspace{0.05in}
\noindent\textbf{UVG Dataset.}
Although \system is specifically designed for representing large-scale and diverse videos, which is not the case for the UVG dataset (7 videos), it can still consistently outperform NeRV greatly as shown in Figure~\ref{fig:uvg_results}.
Specifically, it surpasses NeRV by more than 1.5 dB under the same BPP ratios.
Despite that INR-based and learning-based methods are indeed two different frameworks, we still follow the NeRV paper to compare with learning-based video compression methods for completeness, such as DVC~\cite{lu2019dvc} and DCVC~\cite{li2021deep}.
\system outperforms all of them in both PSNR and MS-SSIM metrics. 
It greatly reveals the effectiveness of our \system for the video compression task.

\begin{table*}[t]
\centering
\begin{minipage}{.42\textwidth}
    \caption{Contribution of each component. SAF, GTMLP, Flow denote spatially-adaptive fusion, global temporal MLP, and multi-scale flow estimation, respectively.}
    \centering
    \resizebox{0.95\linewidth}{!}{
    \renewcommand{\arraystretch}{1.1}
    \begin{tabular}{@{}l|cccc@{}} 
        \toprule
        \multirow{2}{*}{\textbf{Model}} & \multicolumn{2}{c}{\textbf{UVG}} & \multicolumn{2}{c}{\textbf{UCF101}}\\
        \cmidrule(l{-1pt}r{-1pt}){2-5}
         & PSNR & MS-SSIM & PSNR & MS-SSIM \\
        \toprule
        NeRV & 34.13 & 0.948 & 28.00 & 0.935 \\ 
        + GTMLP & 33.94 & 0.946 & 27.96 & 0.935 \\ 
        \midrule
        + SAF & 35.84 & 0.960 & 30.78 & 0.962 \\
        + GTMLP & 36.32 & 0.963 & 30.94 & 0.964 \\
        + Flow & \bf 36.99 & \bf 0.977 & \bf 31.44 & \bf 0.968 \\
        \bottomrule
    \end{tabular}
    }
    \label{tab:components}
\end{minipage}
\hfill
\begin{minipage}{.29\textwidth}
    \caption{Temporal modeling ablation. ``DWConv" indicates the depth-wise temporal conv.}
    \centering
    \resizebox{0.95\linewidth}{!}{
    \renewcommand{\arraystretch}{1.1}
        \begin{tabular}{@{}l|cc@{}} 
            \toprule
            Model & PSNR & MS-SSIM \\
            \toprule
            Baseline & 31.10 & 0.964 \\
            \midrule
            DWConv-k3 & 31.13 & 0.965 \\ 
            DWConv-k7 & 31.15 & 0.966 \\
            DWConv-k11 & 31.16 & 0.966 \\
            Attention & 31.34 & 0.967 \\
            \midrule
            GTMLP & \bf 31.44 & \bf 0.968 \\ 
            \bottomrule
        \end{tabular}
    }
    \label{tab:temporal}
\end{minipage}
\hfill
\begin{minipage}{.24\textwidth}
    \caption{Fusion methods ablation.}
    \centering
    \resizebox{0.9\linewidth}{!}{
    \renewcommand{\arraystretch}{1}
        \begin{tabular}{@{}c|cc@{}} 
            \toprule
            & PSNR & MS-SSIM \\
            \midrule
            U-Net & 30.09 & 0.954 \\ 
            SAF & \bf 31.44 & \bf 0.968 \\
            \bottomrule
        \end{tabular}%
    }
    \label{tab:fusion}
    \vspace{0.08in}
    \caption{Impact of multi-scale.}
    \centering
    \resizebox{0.85\linewidth}{!}{
    \renewcommand{\arraystretch}{1}
        \begin{tabular}{@{}c|cc@{}} 
            \toprule
            MS & PSNR & MS-SSIM \\
            \midrule
             & 31.06 & 0.965 \\ 
            \checkmark & \bf 31.44 & \bf 0.968 \\
            \bottomrule
        \end{tabular}%
    }
    \label{tab:multiscale}
\end{minipage}
\end{table*}

\begin{table*}[t]
\centering
\begin{minipage}{.32\textwidth}
    \caption{Video diversity ablation. We fix the total video count at 1000 while changing the number of action classes.}
    \resizebox{\linewidth}{!}{
    \renewcommand{\arraystretch}{1.15}
    \begin{tabular}{@{}lc|cc@{}} 
        \toprule
        & \textbf{\#Class} & PSNR & MS-SSIM \\
        \midrule
        \multirow{3}{*}{\textbf{NeRV}} 
        & 10 & 27.95 & 0.935 \\
        & 100 & 26.66 & 0.915 \\
        & $\bigtriangledown$ & -1.29 & -0.02 \\
        \midrule
        \multirow{3}{*}{\textbf{\system}} 
        & 10 & 29.74 & 0.950 \\
        & 100 & 29.36 & 0.946 \\ 
        & $\bigtriangledown$ & \bf -0.38 & \bf -0.004 \\
        \bottomrule
    \end{tabular}
    }
    \label{tab:video_diversity}
\end{minipage}
\hfill
\begin{minipage}{.41\textwidth}
    \caption{Top-1 action recognition accuracy on UCF101. Models are trained on compressed videos and tested on uncompressed videos (``Train'' setting) and vice versa (``Test'' setting). S/M/L denote different compression ratios.}
    
    \resizebox{\linewidth}{!}{
    \renewcommand{\arraystretch}{1.1}
    \begin{tabular}{@{}l|ccc|ccc@{}} 
        \toprule
        \multirow{2}{*}{\textbf{Model}} & \multicolumn{3}{c}{\textbf{Train}} & \multicolumn{3}{c}{\textbf{Test}}\\
        \cmidrule(l{-1pt}r{-1pt}){2-7}
         & S & M & L & S & M & L\\
        \toprule
        GT & 91.3 & 91.3 & 91.3 & 91.3 & 91.3 & 91.3 \\ 
        \midrule
        H.264 & 86.7 & 87.9 & 88.9 & 77.2 & 82.4 & 85.5 \\ 
        NeRV & 84.5 & 85.8 & 86.9 & 71.9 & 75.9 & 80.0 \\
        \system & \bf 87.9 & \bf 89.0 & \bf 90.0 & \bf 81.1 & \bf 84.4 & \bf 86.4 \\
        \bottomrule
    \end{tabular}
    }
    \label{tab:ucf_recognition}
\end{minipage}
\hfill
\begin{minipage}{.22\textwidth}
    \caption{Model runtime comparison (video per second).}
    \resizebox{\linewidth}{!}{
    \renewcommand{\arraystretch}{1.1}
    \begin{tabular}{@{}l|c@{}} 
        \toprule
        Method & VPS $\uparrow$ \\
        \midrule
        Frame ({\footnotesize Tab.~\ref{tab:ucf_recognition} GT}) & 273\\ 
        H.264 & 265 \\
        DCVC & 0.9 \\
        NeRV (fp32) & 383 \\
        \system(fp32) & 266  \\
        NeRV (fp16) & 454 \\
        \system(fp16) & 363 \\
        \bottomrule
    \end{tabular}
    }
    \label{tab:runtime}
\end{minipage}
\end{table*}

\subsection{Ablation}

\vspace{0.05in}
\noindent\textbf{Contribution of each component.}
In Table~\ref{tab:components}, we conduct an ablation study to investigate the contribution of each component of \system.
First, we observe that adding the encoder with the spatially-adaptive fusion (SAF) can largely enhance the performance of the baseline model NeRV by 1.7 and 2.8 dB on UVG and UCF101 datasets respectively. 
With the clip-specific visual content fed into the network, it greatly reduces the complexity of memorization for diverse videos.
Second, adding the global temporal MLP module (GTMLP) can further improve performance. It is interesting to note that simply adding the global temporal MLP module on NeRV can not facilitate the final result.
This is because when representing multiple videos, NeRV concatenates all the videos along the time axis. The input of NeRV is the absolute time index normalized by the length of the concatenated video, which can not reflect motion between relative frames.
On the contrary, the input for \system is the relative time index normalized by each video's length, it can represent the motion across frames which are shared across different videos.
Therefore, adding the global temporal MLP module with the relative time index can help model the motion information between frames.
Please note that simply using the relative time index alone is not feasible, which needs to be conditioned on the sampled key-frames to represent different videos. 
Finally, to further reduce the inherent spatial redundancies across video frames, we add the task-oriented flow as an intermediate output, which can boost the final results to another level by 0.67 dB and 0.5 dB on UVG and UCF101 datasets respectively.

\vspace{0.05in}
\noindent\textbf{Component design choices ablation.}
Table~\ref{tab:temporal} demonstrates the results of different temporal modeling designs.
Compared to the baseline, incorporating the local temporal relationship by adding depth-wise temporal convolution can slightly improve the performances and the gap becomes larger while increasing the kernel size from 3 to 11, which validates the importance of temporal modeling for effective video representation.
Inspired by the success of Transformer~\cite{vaswani2017attention}, we also try to add a temporal attention module. Different from convolution operation with the local receptive field, the temporal attention module can model global temporal dependencies, which achieves higher results than depth-wise convolutions. 
However, due to the heavy computation cost of the attention operation, the training speed of the attention module is much slower than other variants.
Finally, motivated by the success of MLP-based models~\cite{tolstikhin2021mlp,touvron2021resmlp,lian2021mlp,he2020gta}, our global temporal MLP module combines the efficiency from fully-connected layer and the global temporal modeling ability from the attention module. It attains the highest results with a much faster training speed than the attention module.
We also compare different fusion strategies to fuse the content information from encoder to decoder in Table~\ref{tab:fusion}.
While U-Net~\cite{ronneberger2015u} concatenates the output feature map of each encoder stage to the input of the decoder, the proposed SAF module utilizes the content feature map as a modulation for decoder features, which proves to be a more effective design than simple concatenation.
In addition, Table~\ref{tab:multiscale} shows that the multi-scale design can enhance the final performances.

\vspace{0.05in}
\noindent\textbf{Impact of video diversity.}
To analyze the impact of video diversity, we conduct experiments with the following settings:
(i) 1000 videos selected from 10 classes where each class has 100 videos;
(ii) 1000 videos selected from 100 classes where each class has 10 videos.
The results are shown in Table~\ref{tab:video_diversity}.
When increasing the video diversity from 10 classes to 100 classes, although the performances of \system and NeRV both decrease, the results of \system drop much slower than NeRV.
It verifies that \system is more effective especially when representing diverse videos.

\subsection{Action Recognition}
\label{sec:recognition}
As \system can effectively represent large and diverse videos, a natural extension of its application could be treating it as an efficient video dataloader, considering it can greatly reduce the video loading time due to the INR-based model design.
In this section, to validate the above assumption, we perform experiments on the action recognition task.

\vspace{0.05in}
\noindent\textbf{Action recognition accuracy.}
In our experiment, we adopt the widely used TSM~\cite{lin2019tsm} as the backbone to evaluate the action recognition accuracy of compressed videos from H.264, NeRV, and \system.
Specifically, we follow two settings below:
i) ``Train'': models are trained on compressed videos and tested on uncompressed ground-truth videos.
ii) ``Test'': models are trained on uncompressed ground-truth videos and tested on compressed videos.
S/M/L denotes different BPP values as Table~\ref{tab:ucf_compression}. The lower BPP value means a higher compression ratio.
The results are shown in Table~\ref{tab:ucf_recognition}.
We can see that, the action recognition accuracy of \system consistently outperforms NeRV by 3-4\% and 6-10\% for the ``train'' and ``test'' settings, respectively.
In addition, \system consistently outperforms H.264, which proves the superior advantage of \system when used as an efficient dataloader in the real-world scenario.

\begin{table*}[t]
\centering
    \caption{Video inpainting comparison between our \system and NeRV. NeRV$^*$ and \system fit all the videos in a shared model. PSNR results of the mask areas are reported here.}
    \resizebox{0.9\linewidth}{!}{
    \renewcommand{\arraystretch}{1.2}
    \begin{tabular}{@{}c|cccccccccc|c@{}} 
        \toprule
        Video & bike & b-swan & bmx & b-dance & camel & c-round & c-shadow & cows & dance-twirl & dog & avg.\\
        \midrule
        NeRV & 19.00 & 21.10 & 18.26 & 18.50 & 18.59 & 16.78 & 19.66 & 18.25 & 17.97 & 21.79 & 18.99\\
        NeRV$^*$ & 20.45 & 21.86 & 19.96 & 19.69 & 20.15 & 18.23 & 20.83 & 18.75 & 18.92 & 22.20 & 19.88 \\
        D-NeRV & 23.53 & 22.27 & 19.50 & 21.85 & 22.89 & 18.9 & 21.06 & 22.27 & 19.08 & 22.01 & \bf 21.3\\
        \bottomrule
    \end{tabular}
    }
    \label{tab:inpaint}
\vspace{-0.05in}
\end{table*}

\vspace{0.05in}
\noindent\textbf{Model runtime.}
In Table~\ref{tab:runtime}, we compare the model runtime of the following settings:
(i) \textbf{Frame}: reading from pre-extracted uncompressed ground-truth frames directly;
(ii) \textbf{H.264}: reading and decoding from the H.264 compressed videos;
(iii) \textbf{DCVC}: recent learning-based video compression method;
(iv) \textbf{NeRV} and (iv) \textbf{\system}.
The experiments are conducted on a single node with 8 RTX 2080ti GPU and 32-core CPU.
Although reading from uncompressed ground-truth frames can preserve the highest quality of video and achieve higher accuracy for downstream tasks, it has a much higher storage cost because it reads from uncompressed frames.
On the contrary, directly reading from compressed videos (\eg, H.264) can save the storage cost while achieving a similar speed because of the highly optimized video decoding techniques.
The model runtime speed of NeRV and \system shows a great advantage over the learning-based compression method DCVC.
Due to its auto-regressive decoding design, DCVC has achieved a much slower model runtime speed than \system and NeRV.
Note that although NeRV has the highest model runtime speed due to the simplicity of its architecture, its compression quality is much inferior to the \system as shown in Table~\ref{tab:ucf_compression} and Table~\ref{tab:ucf_recognition}.

\subsection{Video Inpainting}
\label{sec:inpainting}
We further explore the potential ability of \system on the video inpainting task with NeRV. We apply 5 random box masks with a width of 50 for each frame. The results are shown in Table~\ref{tab:inpaint}. 
Although we do not have any specific design for the video inpainting task, our \system can still outperform NeRV by 1.4 dB for the PSNR results.
Also, it is interesting to see that encoding all videos in a shared model can also improve the inpainting performance (NeRV v.s. NeRV$^{*}$), which further validates our previous claim of encoding all videos in a shared model is more beneficial.

\subsection{Qualitative Results}
\begin{figure}[t]
    \centering
    \adjincludegraphics[width=\linewidth, trim={{0.05\width} {0.13\height} {0.05\width} {0.1\height}},clip]{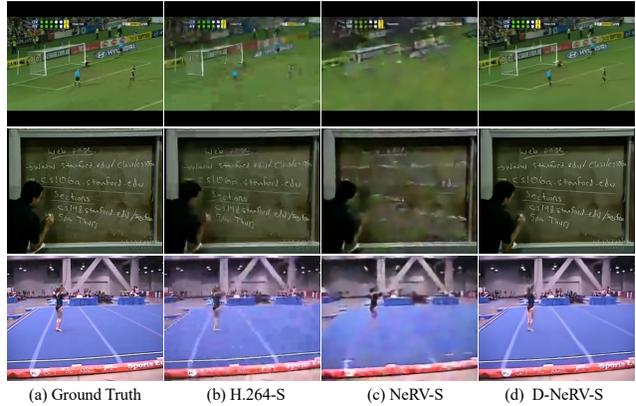}
    \caption{Visualization of ground-truth, H.264, NeRV and \system on the UCF101 dataset. Please zoom in to view the details.}
    \label{fig:qualitative_compression}
\end{figure}

\begin{figure}[t]
    \centering
    \adjincludegraphics[width=\linewidth, trim={{0.05\width} {0.18\height} {0.05\width} {0.18\height}},clip]{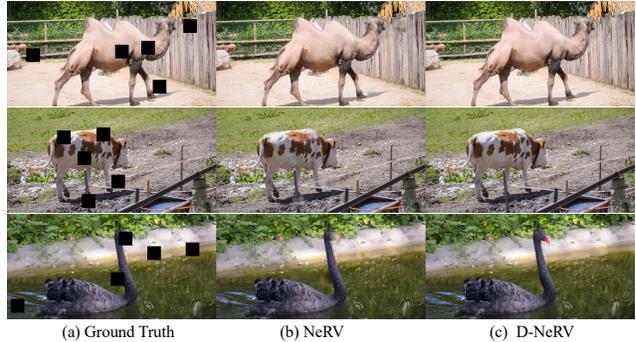}
    \caption{Video inpainting visualization on the DAVIS dataset.}
    \label{fig:qualitative_inpainting}
    \vspace{-0.05in}
\end{figure}

In Figure~\ref{fig:qualitative_compression}, we compare the visualization results of the decoded frames for the compression task. 
At the same BPP budget, \system produces clearer images with higher quality in both the main objects and the background compared to the classic video compression method (H.264) and baselines (NeRV), such as the court, blackboard, and stadium. 
Figure~\ref{fig:qualitative_inpainting} shows the visualization results for the video inpainting task. Compared to NeRV, our \system can inpaint the mask area more naturally with better quality.
More qualitative results are shown in the supplementary material.

%% file: sec/5_conclusions.tex
\section{Conclusion}
In this paper, we present a novel implicit neural network based framework to represent large-scale and diverse videos.
We decouple videos into clip-specific visual content and motion information, and then model them separately, which proves to be more effective than modeling them jointly as previous work NeRV does.
Because it alleviates the difficulty of memorizing diverse videos. 
We also introduce temporal reasoning into the implicit neural network to exploit the temporal relationships across frames.
We further validate our design on multiple datasets and different tasks (\eg, video reconstruction, video compression, action recognition, and video inpainting).
Our method provides new insight into representing videos in a scalable manner, which makes it one step closer to real-world applications.

%% file: appendix.tex
\appendix
\section*{Appendix}

Sec.~\ref{sec:addtional_results} reports additional results on AVA~\cite{gu2018ava} and KTH Action Recognition~\cite{schuldt2004recognizing} datasets for the video compression task.
Sec.~\ref{sec:enerv} presents more comparison with the state-of-the-art INR-based video representation method E-NeRV~\cite{li2022nerv}.
Sec.~\ref{sec:experiment_details} shows more dataset-specific implementation details.
We also show more qualitative results on the UVG, UCF101, Davis datasets in Sec.~\ref{sec:qualitative_more}.
Sec.~\ref{sec:discussion_more} provides more comparisons and discussions with learning-based video compression methods.
Finally, we discuss the limitation and some future work of our paper in Sec.~\ref{sec:limitation}.

\section{Additional Video Compression Results}
\label{sec:addtional_results}
To further demonstrate the effectiveness of \system, we conduct additional experiments on the AVA Actions~\cite{gu2018ava} dataset and KTH Action Recognition~\cite{schuldt2004recognizing} dataset for the video compression task.

\noindent\textbf{AVA Actions Dataset}
For the AVA Actions dataset, each original video is a full movie lasting about 1-2 hours, which is much longer than short action videos (around 10 seconds) from the UCF101 and UVG datasets. 
We sample 10 videos with a spatial size of 256$\times$384 and a frame rate of 1 fps.
The PSNR and MS-SSIM results under different compression ratios (indicated with S / M / L) are shown in Table~\ref{tab:ava}.
We can see that \system consistently outperforms NeRV~\cite{chen2021nerv} and H.264~\cite{wiegand2003overview} when encoding especially long videos.

\noindent\textbf{KTH Action Recognition Dataset}
The KTH Action Recognition~\cite{schuldt2004recognizing} dataset consists of grayscale video sequences of 25 people performing six different actions: walking, jogging, running, boxing, hand waving, and hand clapping. The background is uniform and a single person performs actions in the foreground. The videos have 120$\times$160 spatial size and 25 fps frame rates.
Similar to the results on other datasets, our \system achieves the best performances when comparing to H.264 and NeRV in Table~\ref{tab:kth}.

\section{Additional Comparison with E-NeRV}
\label{sec:enerv}
We conduct an additional comparison with E-NeRV on the UVG dataset by following the same experimental setting as Table \textcolor{red}{1} from E-NeRV~\cite{li2022nerv}.
The original E-NeRV paper uniformly samples 150 frames from each video and resizes the input video from 1080$\times$1920 to 720$\times$1280, and fits each video with a much larger model size (12.5M).
The results of NeRV and E-NeRV in Table~\ref{tab:additional_enerv} are the reported performance in Table \textcolor{red}{1} from the original E-NeRV paper.
As we can see from Table~\ref{tab:additional_enerv}, when using a much larger model size to fit each downsampled video, the PSNR scores of both NeRV and E-NeRV are higher and the performance gap between E-NeRV and NeRV becomes greater, comparing to the results of Table~\textcolor{red}{1} in our main paper.
However, our \system still outperforms E-NeRV by 0.82 dB and achieves the best result.
It proves the superior advantages of \system over the state-of-the-art INR-based video representation method E-NeRV.

\begin{table}[t]
\centering
    \caption{Video compression results on the AVA dataset.}
    \vspace{-0.1in}
    \resizebox{\linewidth}{!}{
    \renewcommand{\arraystretch}{1.1}
    \begin{tabular}{@{}l|ccc|ccc@{}} 
        \toprule
        \multirow{2}{*}{\textbf{Model}} & \multicolumn{3}{c}{\textbf{PSNR}} & \multicolumn{3}{c}{\textbf{MS-SSIM}}\\
        \cmidrule(l{-1pt}r{-1pt}){2-7}
         & S & M & L & S & M & L \\
        \toprule
        H.264 & 27.32 & 28.91 & 30.49 & 0.853 & 0.897 & 0.923 \\ 
        NeRV & 26.48 & 27.28 & 28.21 & 0.840 & 0.868 & 0.893 \\
        \system & \bf 28.77 & \bf 29.57 & \bf 30.60 & \bf 0.886 & \bf 0.903 & \bf 0.924 \\
        \bottomrule
    \end{tabular}
    }
    \label{tab:ava}
\end{table}

\begin{table}[t!]
\centering
    \caption{Video compression results on the KTH dataset.}
    \vspace{-0.1in}
    \resizebox{\linewidth}{!}{
    \renewcommand{\arraystretch}{1.1}
    \begin{tabular}{@{}l|ccc|ccc@{}} 
        \toprule
        \multirow{2}{*}{\textbf{Model}} & \multicolumn{3}{c}{\textbf{PSNR}} & \multicolumn{3}{c}{\textbf{MS-SSIM}}\\
        \cmidrule(l{-1pt}r{-1pt}){2-7}
         & S & M & L & S & M & L \\
        \toprule
        H.264 & 29.61 & 32.72 & 34.51 & 0.691 & 0.801 & 0.860 \\ 
        NeRV & 30.56 & 32.14 & 33.31 & 0.701 & 0.748 & 0.784 \\
        \system & \bf 31.90 & \bf 34.46 & \bf 36.15 & \bf 0.745 & \bf 0.849 & \bf 0.892 \\
        \bottomrule
    \end{tabular}
    }
    \label{tab:kth}
\end{table}

\begin{table*}[t]
    \centering
    \vspace{-0.05in}
    \caption{Video reconstruction comparison between our \system, NeRV~\cite{chen2021nerv} and E-NeRV~\cite{li2022nerv} on 7 videos from the UVG dataset. We follow the same setting as E-NeRV, which uniformly samples 150 frames from each video, resizes the input video from 1080$\times$1920 to 720$\times$1280 and trains models for 300 epochs.}
    \vspace{-0.1in}
    \resizebox{0.75\linewidth}{!}{
    \renewcommand{\arraystretch}{1.25}
    \begin{tabular}{@{}l|ccccccc|c@{}} 
        \toprule
        Video & Beauty & Bosphorus & Bee & Jockey & SetGo & Shake & Yacht & avg.\\
        \midrule
        NeRV & 36.06 & 37.35 & 41.23 & 38.14  & 31.86 & 37.22 & 32.45 & 36.33 \\
        E-NeRV & 36.72 & 40.06 & 41.74 & 39.35 & 34.68 & 39.32 & 35.58 & 38.21 \\
        D-NeRV & 37.53 & 40.74 & 39.89 & 39.94 & 37.51 & 38.85 & 38.63 & \bf 39.03 \\
        \bottomrule
    \end{tabular}
    }
    \label{tab:additional_enerv}
\end{table*}

\begin{figure*}[t!]
    \centering
    \adjincludegraphics[width=\linewidth, trim={{0.13\width} {0.03\height} {0.13\width} {0.02\height}},clip]{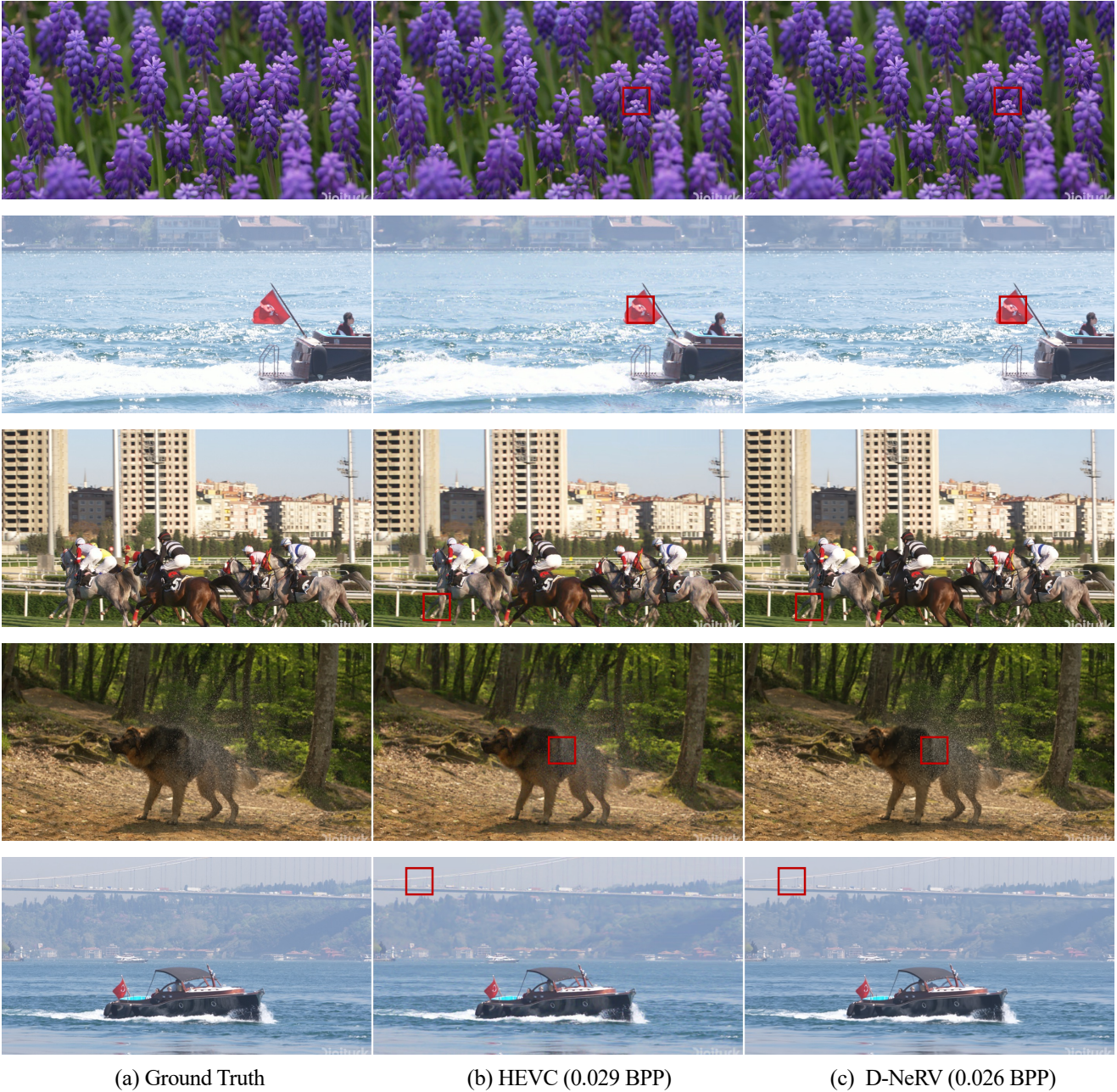}
    \vspace{-0.2in}
    \caption{Visualization of ground-truth, HEVC, and \system for the video compression task on the UVG dataset. 
    Red rectangles highlight the regions that HEVC fails to synthesize correctly and faithfully while our \system succeeds. Please zoom in to see the details.}
    \vspace{-0.2in}
    \label{fig:more_uvg}
\end{figure*}

\begin{figure*}[t]
    \centering
    \vspace{-0.1in}
    \adjincludegraphics[width=\linewidth, trim={{0.03\width} {0.1\height} {0.03\width} {0.1\height}},clip]{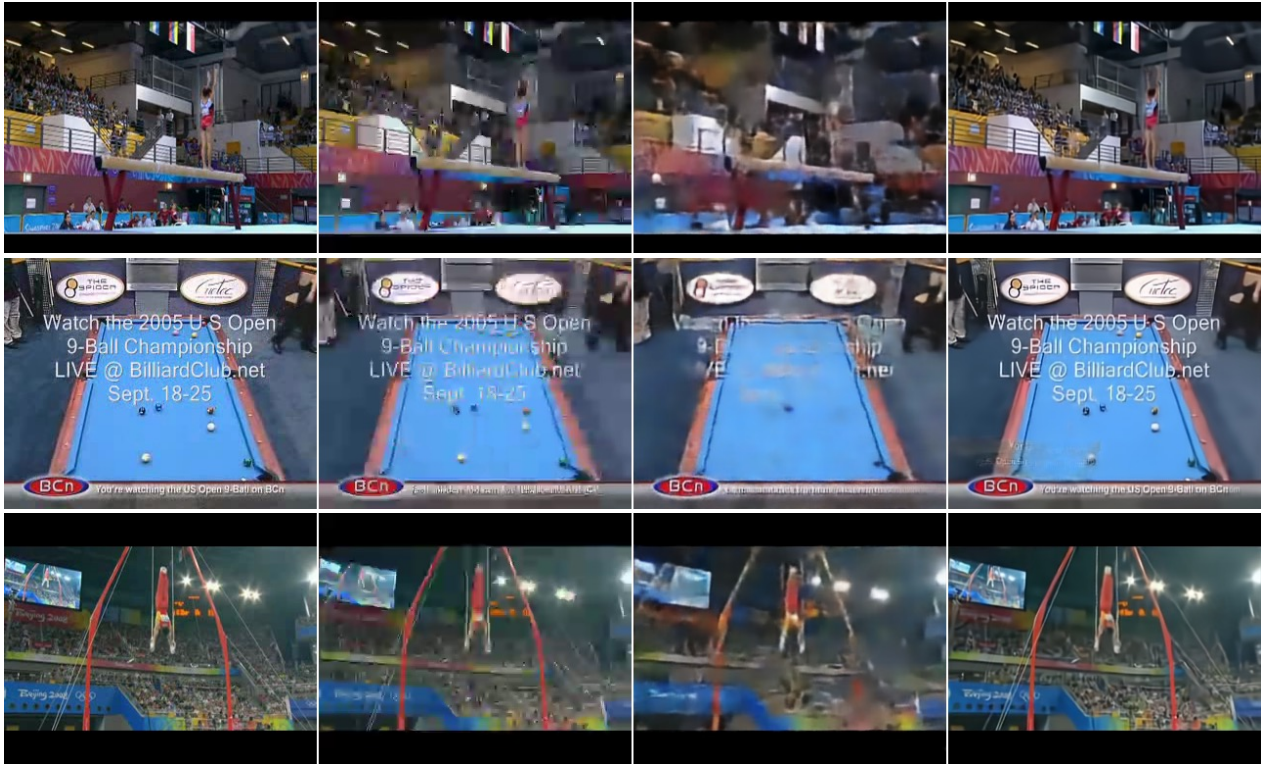}
    \vspace{-0.3in}
    \caption{Visualization of ground-truth, H.264, NeRV and \system for the video compression task on the UCF101 dataset.}
    \label{fig:more_ucf101}
    \vspace{-0.2in}
\end{figure*}

\begin{figure*}[t!]
    \centering
    \vspace{-0.1in}
    \adjincludegraphics[width=\linewidth, trim={{0.03\width} {0.15\height} {0.03\width} {0.1\height}},clip]{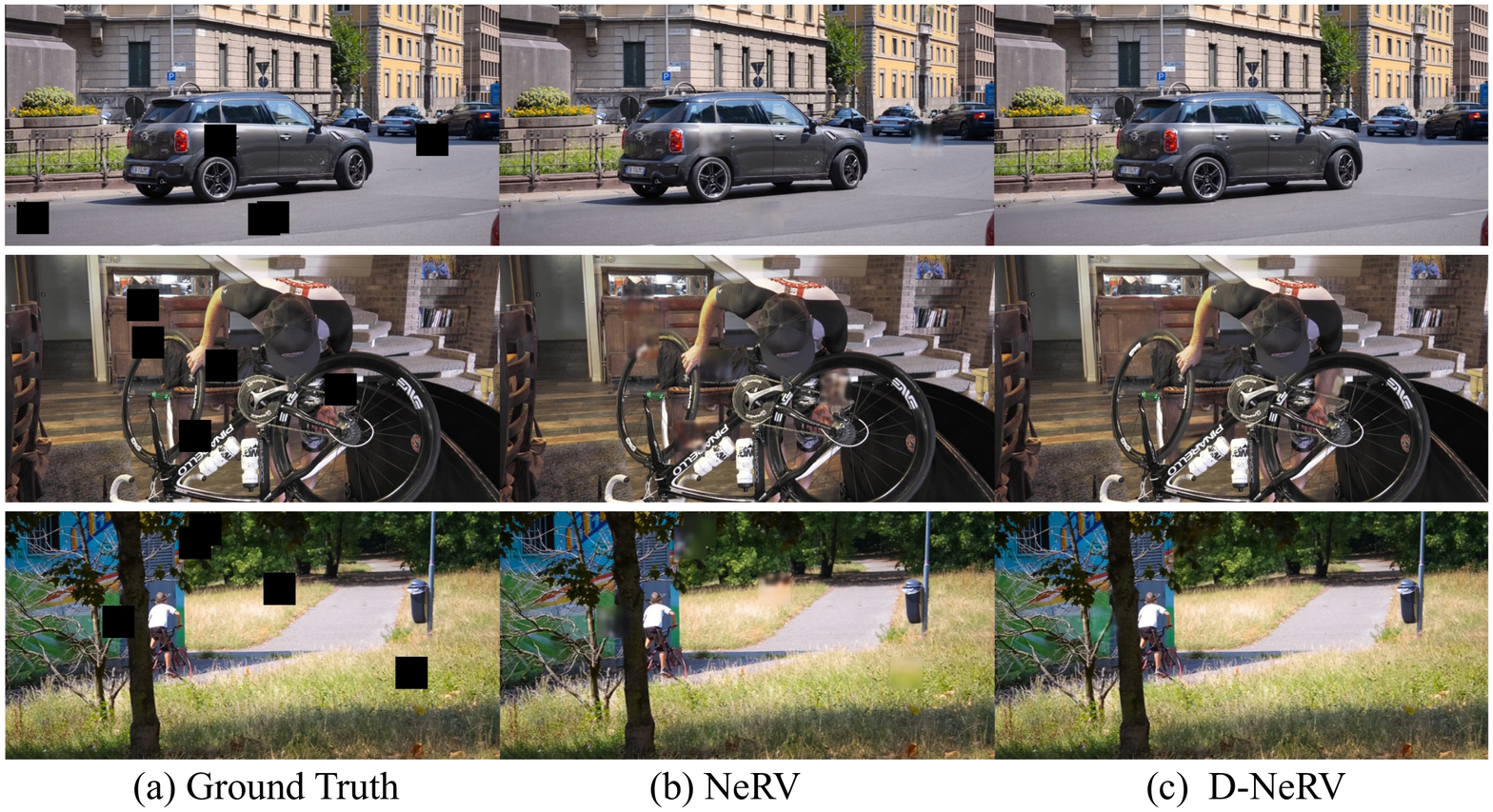}
    \vspace{-0.4in}
    \caption{Visualization of ground-truth, NeRV and \system for the video inpainting task on the Davis dataset. Please zoom in to see the details.}
    \label{fig:more_davis}
    \vspace{-0.2in}
\end{figure*}

\section{Experiment Details}
\label{sec:experiment_details}
On the UVG dataset, to compare with state-of-the-art video compression methods, we run experiments with 1600 epochs and a batch size of 32 and a learning rate of 5e-4. 
Due to the GPU memory limitation, we split the 1024$\times$1920 input video frames into 256$\times$320 image patches. 
We regard the patches at the same spatial location across different timesteps as a single video. 
On the AVA and KTH datasets, we run experiments with 800 epochs, a batch size of 32, and a learning rate of 5e-4.
In our experiments, we set upscale factors 4, 2, 2, 2, 2 for each block.
For the video compression task, following NeRV~\cite{chen2021nerv}, we perform the model quantization and weight encoding steps but without the extra model pruning step to expedite the training process.
And the quantization bit is set to 8 for all the datasets.

For the keyframe image compression, we use pre-trained ~\cite{cheng2020learned} models to compress and decode the keyframes. Different pre-trained image compression models can compress keyframes with different compressed ratios.

\section{More Qualitative Results}
\label{sec:qualitative_more}
\system can produce clearer frames with less noise. Figure~\ref{fig:more_uvg} displays a few samples from the UVG dataset. 
The red rectangles show the regions where our \system outperforms HEVC~\cite{sullivan2012overview}, for example, the flower, the flag, and the leg of the horse.

\system also achieves better qualitative results on UCF101 dataset as shown in Figure~\ref{fig:more_ucf101}. For example, the athlete is more clear than NeRV and H.264 in row 1 and row 3. 
In row 2, \system also distinguishes from other methods when showing the foreground texts.

We also show more qualitative results on the Davis dataset for the video inpainting task in Figure~\ref{fig:more_davis}. \system can inpaint the mask area more faithfully and naturally without blurry effects.

\section{More Discussion}
\label{sec:discussion_more}
In this section, we compare and discuss our \system with existing learning-based video compression methods in more detail. 

The key significant difference between \system and existing learning-based video compression methods is the way compressed videos are represented -- neural network \textit{vs}.\ latent codes, respectively.
Since INR-based \system represents videos as a neural network, it can \textbf{\textit{implicitly}} estimate flow and interpolate keyframes.
In contrast, other learning-based methods that 
 \textbf{\textit{explicitly}} represent videos as latent codes generated by compressing flows and residuals for each frame.
Due to the implicit design, \system is a more general architecture that can be applied to video compression and other video tasks such as video inpainting.
On the other hand, these learning-based methods, including interpolating images (\eg, VCII~\cite{wu2018video}) and predicting flow estimation (\eg, LVC~\cite{rippel2019learned}, SSF~\cite{agustsson2020scale}, FVC~\cite{hu2021fvc}), all decode video frames sequentially because of reliance on previous frames, which leads to a much worse decoding speed.
In contrast, based on INR design, \system can reconstruct video frames parallelly with a faster speed. 

\section{Limitations and Future Work}
\label{sec:limitation}
INR-based methods often require longer training iterations than learning-based compression methods to better capture the high-frequency details. 
And they can not be generalized to unseen videos, which means they can only be trained and tested on the same videos.
We believe more exploration of the generalization ability can be a good research direction for the INR-based video representation models.
In addition, our current \system design still encodes the sampled keyframes by image compression techniques, however, encoding the sampled keyframes by a separate implicit neural network can make the whole pipeline more unified and may achieve better performances.